\PassOptionsToPackage{table}{xcolor}
\documentclass[acmtog, nonacm]{acmart}

\usepackage{booktabs} %

\usepackage[ruled]{algorithm2e} %

\SetAlFnt{\small}
\SetAlCapFnt{\small}
\SetAlCapNameFnt{\small}
\SetAlCapHSkip{0pt}

\usepackage{placeins}       %
\usepackage{booktabs}       %
\usepackage{multirow}       %
\usepackage{tabularx}       %
\usepackage{makecell}       %
\usepackage{array}          %

\newcommand{\method}{ReRoPE}

\AtEndPreamble{
    \usepackage[capitalize]{cleveref}
    \crefname{section}{Sec.}{Secs.}
    \Crefname{section}{Section}{Sections}
    \crefname{table}{Tab.}{Tabs.}
    \Crefname{table}{Table}{Tables}
}

\makeatletter
\DeclareRobustCommand\onedot{\futurelet\@let@token\@onedot}
\def\@onedot{\ifx\@let@token.\else.\null\fi\xspace}

\def\ie{\emph{i.e}\onedot}

\makeatother

\renewcommand\footnotetextcopyrightpermission[1]{}
\begin{document}
\title{ReRoPE: Repurposing RoPE for Relative Camera Control}

\author{Chunyang Li}
\authornote{Equal contribution}
\affiliation{%
  \institution{Zhejiang University}
  \country{China}}

\author{Yuanbo Yang}
\authornotemark[1]
\affiliation{%
  \institution{Zhejiang University}
  \country{China}}

\author{Jiahao Shao}
\authornotemark[1]
\affiliation{%
  \institution{Zhejiang University}
  \country{China}}

\author{Hongyu Zhou}
\affiliation{%
  \institution{Zhejiang University}
  \country{China}}

\author{Katja Schwarz}
\affiliation{%
  \institution{Independent Researcher}
  \country{Germany}}

\author{Yiyi Liao}
\authornote{Corresponding author}
\affiliation{%
  \institution{Zhejiang University}
  \country{China}}

\begin{abstract}
Video generation with controllable camera viewpoints is essential for applications such as interactive content creation, gaming, and simulation. Existing methods typically adapt pre-trained video models using camera poses relative to a fixed reference, e.g., the first frame. However, these encodings lack shift-invariance, often leading to poor generalization and accumulated drift. While relative camera pose embeddings defined between arbitrary view pairs offer a more robust alternative, integrating them into pre-trained video diffusion models without prohibitive training costs or architectural changes remains challenging. We introduce ReRoPE, a plug-and-play framework that incorporates relative camera information into pre-trained video diffusion models without compromising their generation capability. Our approach is based on the insight that Rotary Positional Embeddings (RoPE) in existing models underutilize their full spectral bandwidth, particularly in the low-frequency components. By seamlessly injecting relative camera pose information into these underutilized bands, ReRoPE achieves precise control while preserving strong pre-trained generative priors. We evaluate our method on both image-to-video (I2V) and video-to-video (V2V) tasks in terms of camera control accuracy and visual fidelity. Our results demonstrate that ReRoPE offers a training-efficient path toward controllable, high-fidelity video generation. See project page for more results: \href{https://sisyphe-lee.github.io/ReRoPE/}{\textit{\textcolor{magenta}{https://sisyphe-lee.github.io/ReRoPE/}}}
\end{abstract}



\begin{teaserfigure}
	\includegraphics[width=\textwidth]{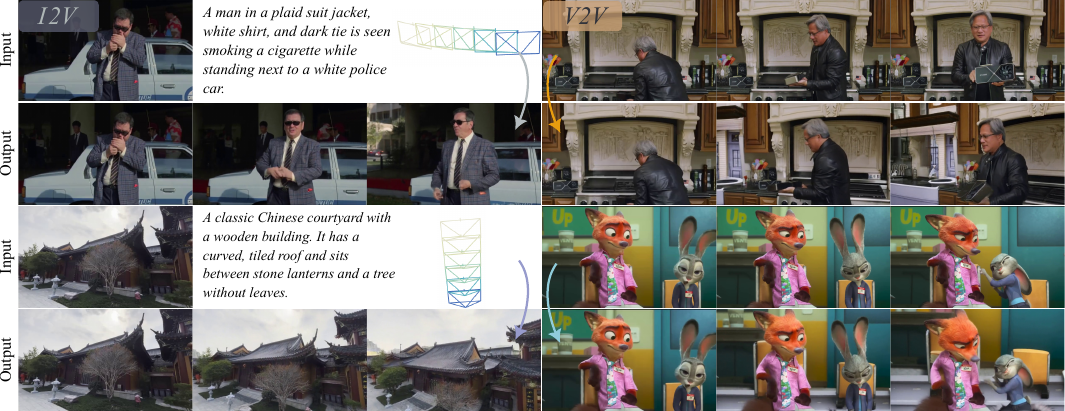}
	\caption{
	\label{fig:teaser}
    ReRoPE enables precise relative camera control for both image-to-video (Left) and video-to-video (Right) tasks by repurposing the redundant low-frequency bands of RoPE.
    }
	\Description[TeaserFigure]{TeaserFigure}
\end{teaserfigure}
\maketitle

\section{Introduction}

Camera control is a fundamental capability desired for video generation models, enabling explicit manipulation of viewpoint while preserving scene identity and temporal consistency. This functionality is essential for applications such as interactive content creation, gaming, and robotics simulation, which require the ability to direct complex cinematic sequences or synthesize diverse perspectives for virtual environments.

Most existing video generation models achieve camera control by injecting 6-DoF poses or ray features through a camera pose encoder~\cite{he2024cameractrl,ren2025gen3c,yang2025das,bian2025gsdit}. These poses define the camera's rotation and translation of all cameras in the trajectory relative to a fixed reference frame, typically the first frame. While this provides explicit viewpoint conditioning, it ties video generation to a specific global reference, limiting generalization to unseen camera trajectories or shifted reference frames. A more robust approach should instead focus on the relative pose change between \textit{any viewpoint pairs} along the trajectory, capturing how the scene appearance evolves based on local camera motion rather than a fixed global coordinate system.
This principle aligns with the design of Rotary Positional Encoding (RoPE)~\cite{su2021roformer,barbero2024round}, which has become the standard in large-scale transformers by shifting from absolute to relative positional representations. This is achieved by applying position-dependent rotations to hidden features, ensuring that the inner product between tokens depends only on their relative \textit{temporal} or \textit{spatial} distance. Since most state-of-the-art video models already employ RoPE, a natural question arises: can we reformulate camera control as a relative signal within this existing relative positional encoding framework?

Prior works have shown that camera parameters can be treated as a form of relative positional encoding similar to RoPE, where 3D transformations between views are injected into the attention mechanism~\cite{gta, li2025prope}. However, these methods typically require training from scratch because they necessitate a fundamental re-partitioning of the RoPE structure. Specifically, RoPE-based video models divide feature vectors into blocks, where each block represents a 1D rotation corresponding to either the temporal, height, or width dimensions. To incorporate camera transformations, existing methods~\cite{li2025prope} often re-allocate these blocks to represent 3D transformations (e.g., $SO(3)$ or $SE(3)$), a configuration that is structurally incompatible with the 1D rotary priors of pre-trained models. Consequently, adopting these relative strategies has historically required re-learning the entire attention mechanism~\cite{gta, li2025prope}. In contrast, training large-scale video diffusion models from scratch is prohibitively expensive, creating a clear demand for methods that can introduce robust camera control while preserving pre-trained architectural priors to ensure rapid convergence and minimal overhead.

\begin{figure}[t]
  \centering
  \includegraphics[width=\linewidth]{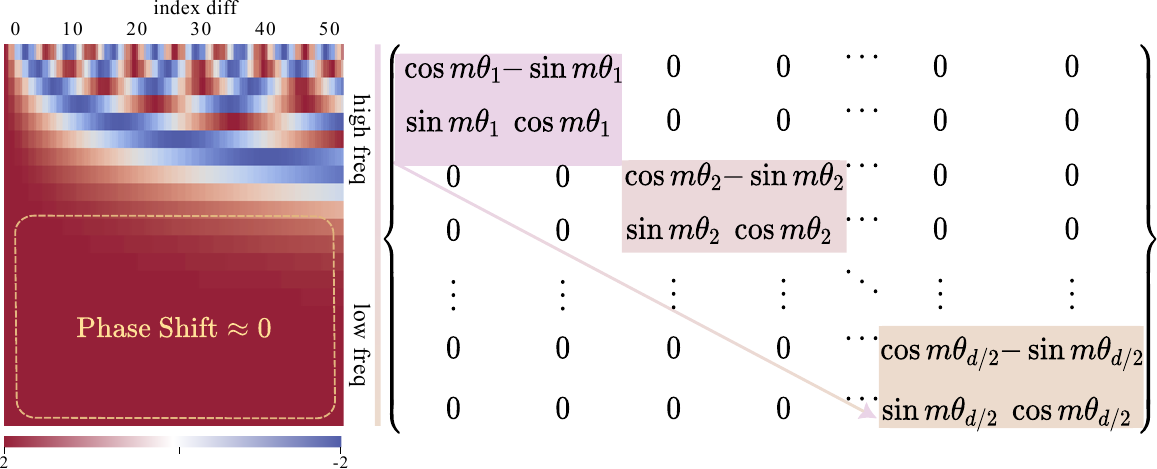}
  \caption{Toy Case Analysis of RoPE. The heatmap shows attention scores of unit key and query vectors across different frequencies and index distances, revealing that low-frequency bands exhibit negligible phase shifts ($\approx 0$) across the window.}
  \label{fig:freq_toy_study}
\end{figure}

We introduce \method, a plug-and-play framework that integrates relative camera information directly into the RoPE modules of a pre-trained video diffusion transformer. Our approach is based on the insight that the RoPE in pre-trained video diffusion models underutilizes its full spectral bandwidth. Specifically, we identify that the low-frequency components of both temporal and spatial RoPE are highly redundant for capturing relative positional differences.
This redundancy is particularly prominent for the temporal dimension, as the temporal sequence length in the latent space of modern video diffusion transformers is typically shorter than the spatial sequence length. Consequently, the temporal RoPE operates primarily in a regime where positional offsets are small, yielding more redundancy compared to the spatial dimensions.
Leveraging this observation, \method~is designed to be intentionally minimal. It requires no modifications to the transformer backbone, no auxiliary camera encoders, and no additional regression heads to interpret camera signals. Instead, relative camera geometry is encoded within a carefully selected subset of these low-frequency temporal channels, leaving the spatial RoPE and high-frequency temporal features untouched. 
This design ensures that \method{} remains fully compatible with existing architectures while strictly preserving pre-trained generative priors. Consequently, our method enables the adaptation of existing models into camera-controllable versions, supporting both image-to-video and video-to-video tasks within a short fine-tuning duration.

Our contributions are three-fold: (1) We identify a consistent low-frequency redundancy within the factorized spatio-temporal RoPE of video transformers, validated through a toy case study and experimental evaluation across three state-of-the-art backbones. (2) Based on this finding, we propose a simple yet effective mechanism to inject relative camera information into these redundant low-frequency bands, achieving high-precision camera control while preserving the pre-trained model's generative priors. (3) Our approach is a plug-and-play module applicable to both video-to-video and image-to-video tasks, where it demonstrates superior camera controllability while maintaining high visual fidelity.

\section{Related Work}
\label{sec:related}

\paragraph{Video Generation Models}
Driven by recent progress in the formulation of generative models like diffusion~\cite{ho2020denoising} and flow matching~\cite{lipman2022flow}, video generation models~\cite{blattmann2023videoldm, blattmann2023stable, wan2021, cogvideo, HaCohen2024LTXVideo, kong2024hunyuanvideo, ali2025world} have achieved remarkable progress.
While earlier methods~\cite{blattmann2023videoldm,blattmann2023stable} adopt U-Net architecture, more recent models~\cite{blattmann2023stable, wan2021, cogvideo, HaCohen2024LTXVideo, kong2024hunyuanvideo}, such as Wan2.1~\cite{wan2021}, switch to Diffusion Transformers (DiT)~\cite{Peebles2022DiT} with full attention mechanisms, achieving impressive performance.
Despite these advances, controllability remains essential for downstream applications. While many recent works~\cite{vace, hu2025hunyuancustom} have achieved diverse control based on pretrained models like style and motion transfer, they still lack the capability for precise camera control.

\paragraph{Camera Control Generation}
Precise camera motion control in generative models follows three primary strategies. Latent-based methods~\cite{guo2023animatediff, sun2024dimensionx, luo2025camclonemaster} utilize weight manipulation (e.g., LoRA) or pattern learning but often lack the explicit geometric constraints required for precise trajectory following. Geometry-guided methods enforce consistency via geometric priors, such as noise warping~\cite{seo2024genwarp} or point cloud projection from depth estimations~\cite{muller2024multidiff, wu2025spmem, yu2024wonderjourney, yu2025wonderworld, duan2025ctrl, huang2025voyager, schneider_hoellein_2025_worldexplorer, yu2024viewcrafter, ren2025gen3c, cao2025uni3c, yu2025trajectorycrafter, bai2025positional}. However, these are strictly limited by depth estimation accuracy, which is non-trivial for dynamic scenes.
The third line of methods bypasses structural estimation by directly injecting camera parameters. These include ray-based approaches~\cite{he2024cameractrl, vanhoorick2024gcd}, which convert 6DoF poses into Plücker ray maps for pixel-wise guidance via adapters, and parameter-based approaches~\cite{wang2024motionctrl, bai2025recammaster, huang2025spacetimepilot}, which fuse encoded extrinsics with visual features. Most such methods define poses relative to a fixed reference (usually the first frame), which is inherently restrictive. We instead propose that relative camera position encodings between \textit{any} two views provide a more scalable and natural integration for controllable video generation.

\paragraph{Relative Positional Encoding.}
Rotary Positional Embedding (RoPE)~\cite{su2021roformer} establishes a standard for incorporating relative positional information in Transformers. By rotating query and key vectors based on their positions, RoPE enforces attention to depend strictly on relative distances. This property has driven its adoption across Large Language Models (LLMs)~\cite{touvron2023llama, dubey2024llama3, qwen2.5, liu2024deepseek} and Vision Transformers (ViTs)~\cite{heo2024ropevit, wang2025vggt, simeoni2025dinov3, flux2024}. 
For video generation, the RoPE mechanism also becomes the default choice for modeling the spatial and temporal positions of the visual tokens~\cite{guo2025lct}. Inspired by RoPE, recent studies~\cite{gta, li2025prope, kong2024eschernet} explore encoding camera relationships through relative embeddings. For instance, PRoPE~\cite{li2025prope} injects camera intrinsics and extrinsics into the self-attention, modeling camera pose as a relative transformation between views. While most of these methods focus on training from scratch to learn the attention with relative camera embedding, concurrent works like BulletTime~\cite{wang2025bullettime} and UCPE~\cite{zhang2025unified} pursue similar directions to ours by injecting relative camera information into the existing RoPE of pre-trained video models.
However, these approaches typically rely on additional learnable weights or auxiliary modules to accommodate the new pose encodings, whereas our method requires no architectural changes based on our findings in the low-frequency redundancy.

\section{Preliminary}
\label{sec:prelim}

\paragraph{Rotary Position Encoding (RoPE)} 
Originally developed for Large Language Models (LLMs) to capture long-range dependencies~\cite{su2021roformer}, RoPE encodes position via feature rotation, ensuring attention depends only on relative token distance. Let $q_i, k_j \in \mathbb{R}^d$ be query and key vectors at positions $i$ and $j$ with head dimension $d$.
RoPE treats the vector as $d/2$ independent 2D planes, rotating each with a frequency $\omega_f = \theta^{-2f/d}$ where $\theta=10^4$ and $f=\{0,\dots,d/2-1\}$. Let  $\mathbf{\Phi}_2(\phi)=\begin{bmatrix}\cos\phi&-\sin\phi\\ \sin\phi&\cos\phi\end{bmatrix}$ denote the $2 \times 2$ rotation, the full rotation matrix for position $m$ is defined as:
\begin{equation}
\mathbf{\Phi}(m)=\mathrm{blkdiag}\!\big(\mathbf{\Phi}_2(m\,\omega_0),\dots,\mathbf{\Phi}_2(m\,\omega_{d/2-1})\big),
\label{eq:rope-1d}
\end{equation}
where $\mathrm{blkdiag}(\mathbf{A}, \mathbf{B}) = \begin{bmatrix} \mathbf{A} & \mathbf{0} \\ \mathbf{0} & \mathbf{B} \end{bmatrix}$ denotes a block-diagonal matrix with $\mathbf{A}$ and $\mathbf{B}$ on its main diagonal, see \cref{fig:freq_toy_study}.
The encoded vectors $\tilde{q}_i = \mathbf{\Phi}(i)\,q_i$ and $\tilde{k}_j = \mathbf{\Phi}(j)\,k_j$ satisfy $\langle \tilde{q}_i, \tilde{k}_j \rangle = \langle q_i, \mathbf{\Phi}(i-j) k_j \rangle$. Thus, attention logits depend on the relative offset $i-j$ rather than absolute indices, providing translation invariance and better length extrapolation.

\paragraph{RoPE for Videos.}
Video diffusion transformers, e.g., Wan~\cite{wan2021} and CogVideoX~\cite{cogvideo}, extend RoPE to the 3D token grid of temporal ($\tau$), height ($h$), and width ($w$) coordinates using a factorized approach. The head dimension $d$ is partitioned into three disjoint channel bands $d = d_\tau + d_h + d_w$, with each band independently encoding a single axis in $(\tau, h, w)$. Let $\mathbf{\Phi}_\tau(\cdot)$, $\mathbf{\Phi}_h(\cdot)$, and $\mathbf{\Phi}_w(\cdot)$ be the rotation operators defined as in \cref{eq:rope-1d} for each band. The 3D rotation operator is the block-diagonal concatenation of these per-axis rotations:
\begin{equation}
\mathbf{\Phi}^{\text{ViT}}(\tau,h,w) = \mathrm{blkdiag} \big( \underbrace{\mathbf{\Phi}_\tau(\tau)}_{d_\tau}, \underbrace{\mathbf{\Phi}_h(h)}_{d_h}, \underbrace{\mathbf{\Phi}_w(w)}_{d_w} \big),
\label{eq:rope-3d}
\end{equation}
where $d_\tau=d_h=d_w=d/3$ is a common choice for popular video diffusion models~\cite{cogvideo, wan2021}
By the orthogonality of the block-diagonal structure, the inner product between two tokens at coordinates $(\tau, h, w)$ and $(\tau', h', w')$ depends solely on their relative spatio-temporal offsets $(\Delta \tau, \Delta h, \Delta w)$.

\paragraph{Cameras as Relative Positional Encoding.}

Similar to RoPE for videos, GTA~\cite{gta} and PRoPE~\cite{li2025prope} adopt a factorized construction of the rotation matrix $\mathbf{\Phi}$ for viewpoint interpolation between posed input frames. Specifically, they allocate a disjoint band of the head dimension to represent relative camera transformations. The remaining channels encode the relative spatial location $(h, w)$. Let $\mathbf{P}_c = \mathbf{K}_c [\mathbf{R}_c \mid \mathbf{t}_c]$ denote the $3 \times 4$ projection matrix of a camera $c$, which is lifted to a $4 \times 4$ homogeneous matrix:
\begin{equation}
\tilde{\mathbf{P}}_c=
\begin{bmatrix}
\mathbf{K}_c\mathbf{R}_c & \mathbf{K}_c\mathbf{t}_c\\
\mathbf{0}^\top & 1
\end{bmatrix}.
\label{eq:lift-P}
\end{equation}
For a token at camera position $\tilde{\mathbf{P}}_c$, the camera information is injected via a block-diagonal transformation $\mathbf{\Phi}^{\text{PRoPE}}(c, h, w) \in \mathbb{R}^{d \times d}$:
\begin{equation}
\mathbf{\Phi}^{\text{PRoPE}}(c,h,w) = \mathrm{blkdiag} \big( \underbrace{\mathbf{\Phi}_{\text{proj}}(c)}_{d_c}, \underbrace{\mathbf{\Phi}_h(h)}_{d_h}, \underbrace{\mathbf{\Phi}_w(w)}_{d_w} \big)
\label{eq:prope-Dproj}
\end{equation}
where $d_c=d/2$ and $d_h=d_w=d/4$ in PRoPE~\cite{li2025prope}. Here, $\mathbf{\Phi}_{\text{proj}}(c)$ is constructed by repeating the lifted camera matrix $\tilde{\mathbf{P}}_c$ along the diagonal:
\begin{equation}
\mathbf{\Phi}_{\text{proj}}(c) = \mathrm{blkdiag} \big( \tilde{\mathbf{P}}_c, \dots, \tilde{\mathbf{P}}_c \big) \in \mathbb{R}^{d_c \times d_c}.
\end{equation}
and $\tilde{\mathbf{P}}_c$ is the lifted matrix for the camera at index $c$. By partitioning the head dimension this way, the attention mechanism computes the relative transformation between any two tokens at positions $c$ and $t$:
\begin{equation}
\mathcal{P}_{c\leftarrow t} = \tilde{\mathbf{P}}_{c} \tilde{\mathbf{P}}_{t}^{-1}.
\label{eq:rel-proj}
\end{equation}
Since $\mathcal{P}_{c\leftarrow t}$ depends only on the camera pair, it encodes a signal suitable for relative positional modeling. However, as can be observed, the spatial-temporal RoPE design in \cref{eq:rope-3d} and the camera-projection design in \cref{eq:prope-Dproj} utilize different channel-wise structures. This architectural discrepancy makes it difficult to directly merge relative camera pose embeddings into pre-trained video models without disrupting their generative priors.

\section{\method}
\label{sec:motivation}

\subsection{Overview and setting.}
Our goal is to enable precise camera control within a pre-trained video diffusion transformer. We consider two distinct generative settings: 1) Video-to-Video (V2V): Given an input video sequence and its corresponding source camera trajectory, the task is to re-render or adapt the video content to follow a new, user-defined target camera trajectory. 2) Image-to-Video (I2V): Given a single reference frame, the model synthesizes a video sequence that adheres to a specified target camera trajectory. 

We build upon a pre-trained video diffusion transformer operating on a 3D latent grid of $T \times H \times W$ tokens. The model employs a multi-head attention mechanism with head dimension $d = d_\tau + d_h + d_w$, split into factorized temporal, vertical, and horizontal RoPE bands. We denote the temporal index by $\tau \in \{0, \dots, T-1\}$ and the spatial indices by $(h, w)$. 

\subsection{Low-Frequency Redundancy of RoPE}
As discussed in \cref{sec:prelim}, it is non-trivial to inject the relative camera pose into the existing RoPE matrix without disrupting the original generative video prior. Therefore, we first look into the original RoPE and analyze its behavior.

\paragraph{Toy Case of Frequency Bands} 
Inspired by~\cite{barbero2024round}, which demonstrates that attention mechanisms of LLMs often use only a subset of RoPE frequency bands, we investigate the specific impact of different bands on video generation. 

We design a toy experiment.
Consider two-dimensional query and key tokens at temporal positions $i$ and $j$. To isolate the effect of the rotation, we set all features to a constant unit vector $q_i = k_j = [1, 1]^\top$. The resulting attention score $A$ after the RoPE operation $\mathbf{\Phi}_2$ is defined by the inner product:
\begin{equation} 
\begin{aligned} 
A(i, j, \omega_k) &= \langle \mathbf{\Phi}_2(i \omega_k) q_i, \mathbf{\Phi}_2(j \omega_k) k_j \rangle \\ &= \langle q_i, \mathbf{\Phi}_2((i-j) \omega_k) k_j \rangle = 2 \cos((i-j) \omega_k),
\end{aligned} 
\end{equation}
where $\omega_k = \theta^{-2f/d}$ with $\theta=10^4$ and $f=\{0,\dots,d/2-1\}$. 
For our experiment, we use common values in video transformers, considering $50$ frames, \ie, $i,j\in\{0,...,49\}$, and a base frequency of $\theta=10^4$.
We visualize frequency bands indexed from high to low frequency on the x-axis and the relative distance between temporal positions $\Delta \tau = i - j$ on the y-axis in \cref{fig:freq_toy_study} (left).

The resulting heatmap reveals an oscillatory attention signal where high-frequency bands (small $f$) shift phase rapidly, providing fine-grained positional discrimination. Conversely, low-frequency bands (large $f$) remain nearly constant across standard video lengths, with values approaching the dot product $\langle [1,1]^\top, [1,1]^\top\rangle = 2$. This indicates that these low-frequency components are underutilized for positional encoding in pre-trained models. This redundancy suggests that the low-frequency subspace of $\mathbf{\Phi}_\tau$ can be repurposed for camera geometry without disrupting the temporal ordering captured by higher frequencies.

\begin{figure}[t]
  \centering
  \includegraphics[width=\linewidth]{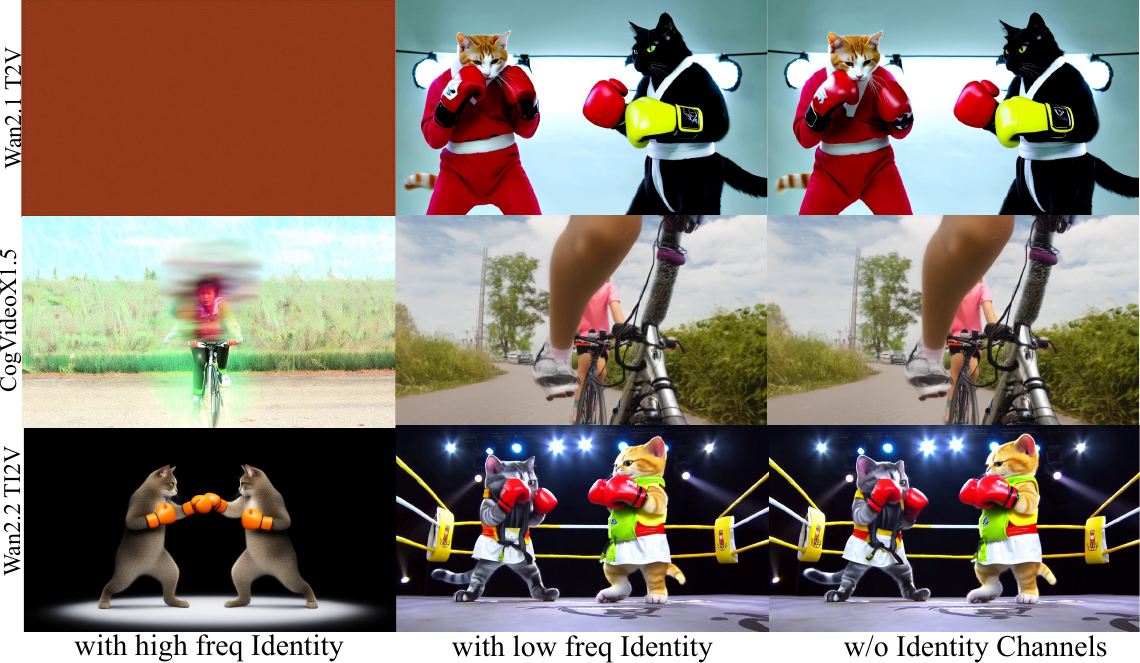}
  \vspace{-0.4cm}
  \caption{ Frequency analysis across three video diffusion models~\cite{kong2024hunyuanvideo, cogvideo, wan2021}. We observe that masking high-frequency bands (Left) leads to model collapse, whereas masking low-frequency bands (Middle) maintains generation quality comparable to the baseline (Right), confirming the functional redundancy of low-frequency components.}
  \label{fig:motivation}
  \vspace{-0.2cm}
\end{figure}

\paragraph{Disabling Frequency Bands in Video DiTs}
To further examine the effect of different frequency bands in trained video diffusion models, we investigate the behavior of ``masking out'' low-frequency bands on three well-known models: Wan 2.1~\cite{wan2021}, Wan 2.1-I2V, and CogVideoX~\cite{cogvideo}.
We set the rotary matrix to the identity for the low-frequency subspace corresponding to $f \in \{d_\tau/4, \dots, d_\tau/2-1\}$:
\begin{equation}
\mathbf{\Phi}(m)=\mathrm{blkdiag}\!\big(
\underbrace{\mathbf{\Phi}_2(m\,\omega_0),\dots,\mathbf{\Phi}_2(m\,\omega_{d_\tau/4-1})}_{d_\tau/2},
\underbrace{\mathbf{I}}_{d_\tau/2}
\big),
\label{eq:rope-1d}
\end{equation}
This masking is applied to all spatial and temporal RoPE components, i.e., to $\mathbf{\Phi}_\tau(\tau)$, $\mathbf{\Phi}_h(h)$, and $\mathbf{\Phi}_w(w)$.
As shown in \cref{fig:motivation}, the videos generated with this ``masking'' remain nearly identical to their original counterparts, indicating that the models do not rely on these bands for discriminating relative positions. In contrast, masking out the high-frequency bands ($f \in \{0, \dots, d/4-1\}$) significantly degrades the model's performance. This empirical result further corroborates our key insight that a substantial portion of the RoPE head dimension is functionally redundant and can be repurposed for external control signals like camera geometry.

\begin{figure*}[t]
  \centering
  \includegraphics[width=0.95\textwidth]{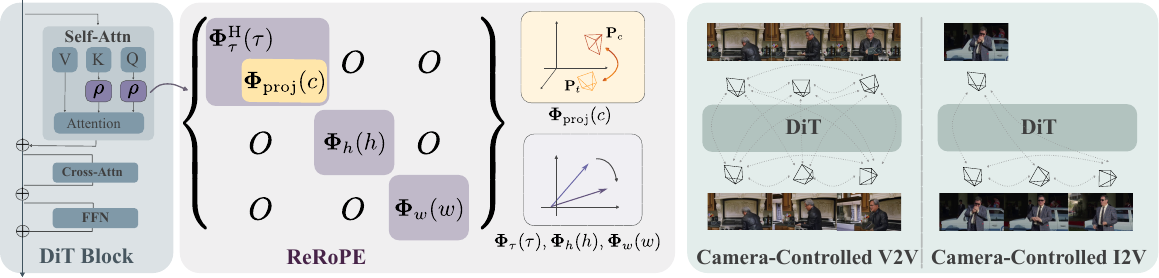}
  \vspace{-0.3cm}
  \caption{Overview of \method{} framework. We enable relative camera control by repurposing the redundant low-frequency temporal bands of a pre-trained Video DiT. (Left and Middle) As shown in the matrix decomposition, we retain spatial and high-frequency temporal bands to preserve generative priors, replacing only the underutilized low-frequency temporal bands with our camera projection block. (Right) This plug-and-play mechanism supports both Video-to-Video and Image-to-Video tasks.}
  \Description{Pipeline diagram showing how ReRoPE injects camera information into the low-frequency temporal RoPE bands while preserving high-frequency temporal and spatial RoPE structure.}
  \vspace{-0.2cm}
  \label{fig:pipeline}
\end{figure*}

\subsection{ReRoPE: Repurposing RoPE for Camera Control}
\label{sec:inject}
We now repurpose the temporal low-frequency subspace to carry camera information,
while preserving the original RoPE behavior on all remaining channels, as shown in \cref{fig:pipeline}. 
While our analysis also identifies redundancy in the spatial embeddings, we focus on applying ReRoPE to the temporal dimension, because the (latent) number of frames is typically smaller than the (latent) spatial resolution. E.g., $T=21, H=30, W=52$ for Wan2.1~\cite{wan2021}, meaning that the temporal embeddings contain the highest redundancy.

\paragraph{ReRoPE at Temporal RoPE}
Our proposed solution is simple: we keep the standard temporal RoPE on the high-frequency bands, and replace the low-frequency band with relative camera information.
For the relative camera, we follow PRoPE to construct the projection block as shown in Eq.~\ref{eq:prope-Dproj}).
Let $d_L$ denote the dimensionality of the temporal low-frequency subspace we choose to repurpose, which is a value that is divisible by $4$.
The repurposed temporal operator is

\begin{equation}
\mathbf{\Phi}^{\text{ReRoPE}}(\tau,c,h,w) = \mathrm{blkdiag} \big( \underbrace{\mathbf{\Phi}^{\text{H}}_\tau(\tau)}_{d_\tau^H},
\underbrace{\mathbf{\Phi}_{\text{proj}}(c)}_{d_\tau^L}, \underbrace{\mathbf{\Phi}_h(h)}_{d_h}, \underbrace{\mathbf{\Phi}_w(w)}_{d_w} \big),
\label{eq:rope-3d}
\end{equation}
i.e., we only modify the temporal low-frequency bands and keep all spatial RoPE bands intact. Here, $d_\tau^H=d_\tau^L=d/6$ in our experiments.
Note that in contrast to the original RoPE operations, the relative camera encoding operation achieved via $\mathbf{\Phi}_\text{proj}(c)$ encodes the relative camera transformation  $\mathcal{P}_{c\leftarrow t} = \tilde{\mathbf{P}}_{c} \tilde{\mathbf{P}}_{t}^{-1}$ between any pair of cameras $c$ and $t$, which is non-norm-preserving.
Therefore, we stabilize training by normalizing the translation component when constructing $\tilde{\mathbf{P}}_c$ and $\tilde{\mathbf{P}}_t$, as detailed in ~\cref{sec:training}.

\subsection{Training}
\label{sec:training}

We fine-tune the pre-trained video diffusion transformer to incorporate explicit camera control. We train our models for Video-to-Video (V2V) and Image-to-Video (I2V) settings, respectively.

\begin{figure*}[t]
  \centering
  \includegraphics[width=\textwidth]{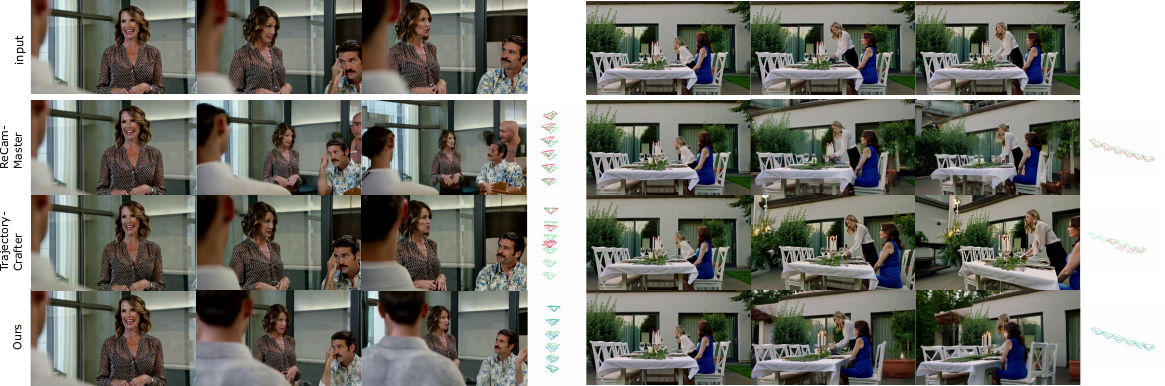}
  \vspace{-0.8cm}
  \caption{
  Qualitative comparison of V2V generation. Trajectory visualization shows that baselines (red) deviate from the ground truth (green). In contrast, \method{} (blue) maintains tight alignment, demonstrating the precise geometric control required for superior visual fidelity and consistency.
  }
  \Description{Qualitative comparison between ReRoPE and baseline methods showing video frames with different camera trajectories.}
  \label{fig:baseline_v2v}
  \vspace{-0.2cm}
\end{figure*}

\begin{table*}[t]
\centering
\scriptsize
\setlength{\tabcolsep}{2.6pt}
\renewcommand{\arraystretch}{1.18}
\begin{tabularx}{\textwidth}{l *{10}{>{\centering\arraybackslash}X}}
\toprule
Method &
\multicolumn{3}{c}{Camera Accuracy} &
\multicolumn{6}{c}{VBench} &
View \\
\cmidrule(lr){2-4}\cmidrule(lr){5-10}\cmidrule(lr){11-11}
& RRE$\downarrow$ & RTE$\downarrow$ & ATE$\downarrow$
& \makecell{subject\\consistency$\uparrow$}
& \makecell{background\\consistency$\uparrow$}
& \makecell{motion\\smoothness$\uparrow$}
& \makecell{dynamic\\degree$\uparrow$}
& \makecell{aesthetic\\quality$\uparrow$}
& \makecell{imaging\\quality$\uparrow$}
& View Syn$\uparrow$ \\
\midrule
TrajectoryCrafter~\cite{yu2025trajectorycrafter}    &   1.1693   & 0.0741 & 0.3824 &    \textbf{0.9406} & \textbf{0.9549} & 0.9842  & 0.4426 & \textbf{0.6176} & 0.6174 & 1147    \\
ReCamMaster~\cite{bai2025recammaster} &
0.7758 & 0.2434 & 0.4814 &
0.9270 & 0.9333 & \textbf{0.9904} & 0.8985 & 0.5831 & 0.6497 & 1481 \\
\rowcolor{gray!12}
ReRoPE (Ours) &
\textbf{0.7416} & \textbf{0.0629} & \textbf{0.1853} &
0.9258 & 0.9319 & \textbf{0.9904} & \textbf{0.9045} & 0.5824 & \textbf{0.6583} & \textbf{1600} \\
\bottomrule
\end{tabularx}
\caption{V2V camera control on SDG-1.5M~\cite{huang2025vipe} (200 sampled videos). We report camera pose errors (RRE/RTE/ATE), VBench sub-metrics, and View Syn. $\uparrow$ higher is better, $\downarrow$ lower is better.}
\vspace{-0.3cm}
\label{tab:v2v_sdg}
\end{table*}

\paragraph{Conditioning Strategy.}
In the V2V setting, we treat generation as conditional video inpainting. We concatenate the latent representations of the source and target videos along the temporal dimension, applying the diffusion process exclusively to target frames. This setup forces the model to learn geometric correlations across all source-source, source-target, and target-target view pairs based on their relative camera trajectories.
For I2V, the first frame serves as a noise-free visual anchor, and the model predicts the noise for the subsequent sequence. To support Classifier-Free Guidance (CFG)~\cite{ho2022classifier} during inference, we apply a 10\% dropout rate to text prompts during training.

\paragraph{Relative Pose Normalization}
To ensure relative camera transformations are meaningful across all view pairs, we define source and target poses within a unified coordinate system. Specifically, in V2V, the first frame of the condition sequence serves as the origin for both trajectories, while in I2V, poses are relative to the input image frame. To resolve scale ambiguity across diverse scenes, we normalize all translation vectors by the maximum $L_2$ norm of translations within the training sequence. This yields a scale-invariant pose representation compatible with our ReRoPE injection.

\paragraph{Objective and Optimization}
We freeze all parameters of the pre-trained backbone except for the self-attention layers, allowing the model to adapt to the new positional embedding mechanism without forgetting its generative prior.
The model is optimized using the flow matching objective~\cite{lipman2022flow}, computed as the Mean Squared Error (MSE) loss between the predicted and ground-truth flow velocity.
We use the AdamW optimizer with a learning rate of $1\mathrm{e}{-5}$ and gradient clipping of $0.05$.

\section{Experiments}
\label{sec:experiments}

We evaluate \method{} on camera-controlled video generation.
We first describe the experimental setup in \S\ref{sec:exp-setup}, then compare against state-of-the-art baselines on camera controllable video-to-video (V2V) and image-to-video (I2V) settings (\S\ref{sec:v2v}–\S\ref{sec:i2v}). Finally, we present ablations on key design choices in \S\ref{sec:ablation}.
\subsection{Experimental Setup}
\label{sec:exp-setup}

We employ Wan2.1 and Wan2.2 as our base video diffusion transformers for the V2V and I2V tasks, respectively, modifying only the temporal low-frequency RoPE bands while keeping the denoising architecture intact. For V2V, we train on the calibrated MultiCamVideo~\cite{bai2025recammaster} dataset; for I2V, we jointly train on MultiCamVideo, SDG-1.5M~\cite{huang2025vipe} (annotated via VIPE~\cite{huang2025vipe}), and DL3DV~\cite{ling2024dl3dv}. The models are trained for 10,000 iterations on 8 NVIDIA RTX 5880 GPUs, requiring approximately one day for V2V and 2$\sim$3 days for I2V. Performance is evaluated using: (i) camera accuracy via pose errors (RRE/RTE/ATE) estimated by VIPE, (ii) video quality and consistency through VBench~\cite{huang2023vbench} sub-metrics, and (iii) geometric fidelity via View Synchronization (View Syn). View Syn uses GIM~\cite{shen2024gim} to calculate the average number of matched pixel pairs per frame with high confidence between generated and ground truth frames. For our ablation studies, we additionally report FID~\cite{heusel2017gans} and FVD~\cite{unterthiner2018towards}.

\subsection{V2V: Comparison to Baselines}
\label{sec:v2v}
\begin{figure*}[t]
  \centering
  \includegraphics[width=\textwidth]{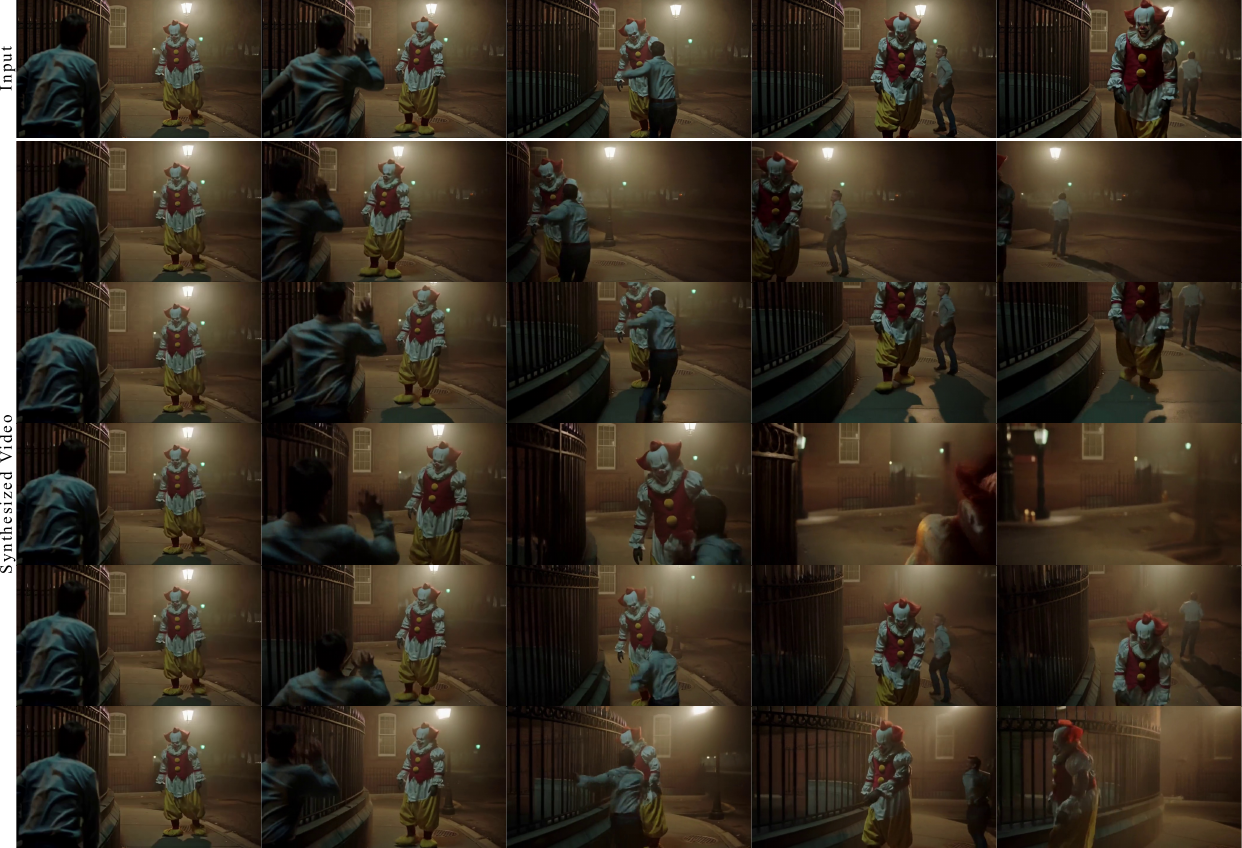}
  \caption{More results of camera-controlled V2V, with each row displaying a distinct camera trajectory. }
  \Description{Qualitative comparison between ReRoPE and baseline methods showing video frames with different camera trajectories.}
  \label{fig:more_results_v2v}
\end{figure*}

We evaluate video-to-video (V2V) camera control on 200 videos sampled from SDG-1.5M. All methods use identical conditioning videos, target trajectories, prompts, and seeds to ensure a fair comparison, and none of them has been trained on SDG-1.5M for the V2V setting. We compare ReRoPE against two state-of-the-art baselines, TrajectoryCrafter~\cite{yu2025trajectorycrafter} and ReCamMaster~\cite{bai2025recammaster}, as shown in Table~\ref{tab:v2v_sdg} and \cref{fig:baseline_v2v}.
TrajectoryCrafter relies on estimating the depth of the input to construct a dynamic point cloud, followed by explicit reprojection to achieve camera control. While this approach achieves superior aesthetic quality, it is highly sensitive to depth estimation errors, and inaccuracies in the global point cloud lead to lower camera accuracy. In contrast, \method\ avoids explicit geometric conditioning, reflecting in higher camera accuracy.
ReCamMaster is more closely related to our framework, as it also avoids explicit geometry conditioning by projecting camera parameters through an encoder and adding them directly to the transformer's intermediate features. However, our results show that \method{} achieves superior camera controllability by encoding the relative relationship between any pair of cameras within the RoPE subspace while maintaining competitive perceptual quality. \cref{fig:more_results_v2v} further illustrates our V2V results under various camera trajectories.

\begin{figure*}[t]
  \centering
  \includegraphics[width=\textwidth]{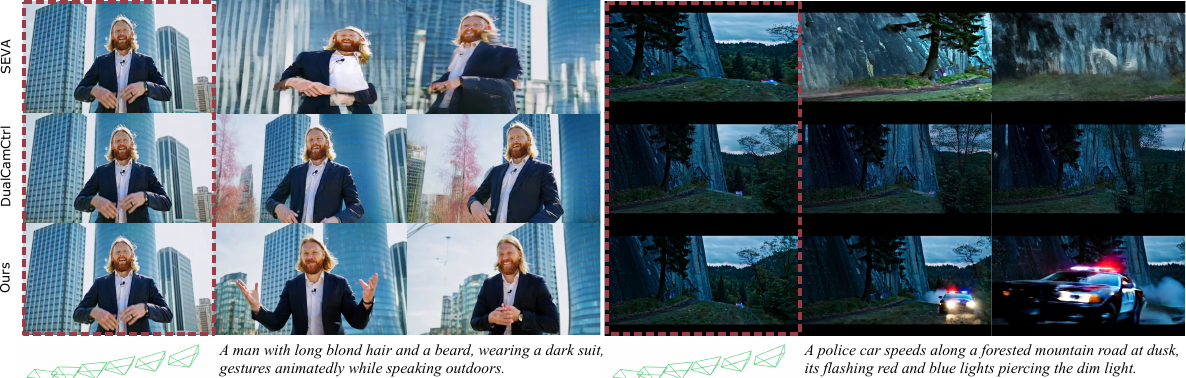}
  \vspace{-0.4cm}
  \caption{Qualitative comparison on dynamic I2V scenes showing that ReRoPE synthesizes natural object motion, whereas baselines incorrectly render the subject as a static rigid body. }
  \label{fig:baseline_i2v}
  \vspace{-0.2cm}
\end{figure*}

\begin{table*}[t]
\centering
\scriptsize
\setlength{\tabcolsep}{2.2pt}
\renewcommand{\arraystretch}{1.18}
\begin{tabularx}{\textwidth}{l *{9}{>{\centering\arraybackslash}X}}
\toprule
Method &
\multicolumn{3}{c}{Camera Accuracy} &
\multicolumn{6}{c}{VBench} \\
\cmidrule(lr){2-4}\cmidrule(lr){5-10}
& RRE$\downarrow$ & RTE$\downarrow$ & ATE$\downarrow$
& \makecell{subject\\consistency$\uparrow$}
& \makecell{background\\consistency$\uparrow$}
& \makecell{motion\\smoothness$\uparrow$}
& \makecell{dynamic\\degree$\uparrow$}
& \makecell{aesthetic\\quality$\uparrow$}
& \makecell{imaging\\quality$\uparrow$} \\
\midrule

SEVA~\cite{zhou2025stable} & 
 0.2767 & 0.0642 &  0.3183&
0.9334 & 0.9445 & 0.9766 & 0.2980 & \textbf{0.5431} & 0.6384 \\

DualCamCtrl~\cite{zhang2025dualcamctrl} & 
 0.1080 & 0.0103 & 0.0909 &
0.9543 & \textbf{0.9462} & 0.9846 & - & 0.5163 & \textbf{0.7084} \\

\rowcolor{gray!12}
ReRoPE (Ours) & 
\textbf{0.0886} &  \textbf{0.0078} & \textbf{0.0703} &
 \textbf{0.9638} & 0.9403 & \textbf{0.9896} & \textbf{0.5587} & 0.5382 & 0.7020 \\
\bottomrule
\end{tabularx}
\caption{I2V camera control evaluated on a test set composed of  \textbf{DL3DV}. We report camera pose errors (RRE/RTE/ATE) and VBench sub-metrics. $\uparrow$ higher is better, $\downarrow$ lower is better.}
\label{tab:i2v_mix}
\vspace{-0.2cm}
\end{table*}

\subsection{I2V: Comparison to Baselines}
\label{sec:i2v}
\begin{figure*}[t]
  \centering
  \includegraphics[width=\textwidth]{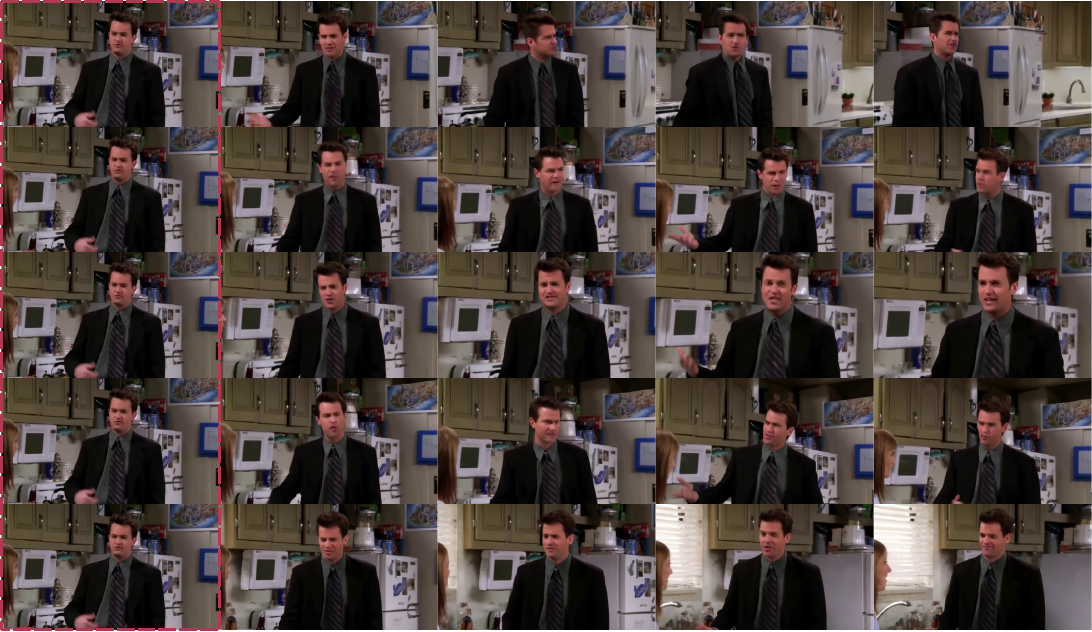}
  \caption{More results of camera-controlled I2V,  with each row displaying a distinct camera trajectory.}
  \Description{Qualitative comparison between ReRoPE and baseline methods showing video frames with different camera trajectories.}
  \label{fig:more_results_i2v}
\end{figure*}

\begin{figure*}[t]
  \centering
  \includegraphics[width=\textwidth]{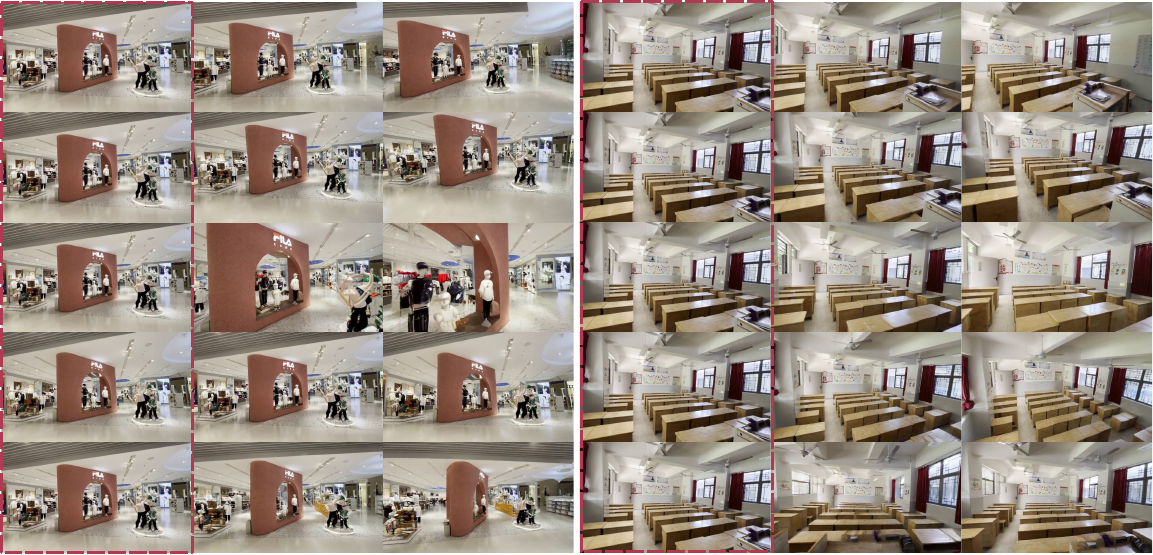}
  \caption{More results of camera-controlled I2V on DL3DV. }
  \Description{Qualitative comparison between ReRoPE and baseline methods showing video frames with different camera trajectories.}
  \label{fig:more_results_dv3dv}
\end{figure*}

We evaluate \method{} for I2V camera control on DL3DV~\cite{ling2024dl3dv} (static scenes) and SDG-1.5M~\cite{huang2025vipe} (featuring dynamic subjects). A key advantage of our approach is the ability to simultaneously model camera motion and object dynamics. In contrast, existing baselines are largely limited to static viewpoint transfer, where dynamic subjects appear as rigid "frozen" scans. Because pose estimation algorithms struggle to decouple camera movement from scene motion in dynamic videos, quantitative metrics become unreliable for comparing rigid vs. non-rigid outputs. Therefore, we restrict our quantitative comparison to the static scenes of DL3DV to ensure a fair evaluation of camera accuracy.
Specifically, we compare our model with two state-of-the-art baselines: SEVA~\cite{zhou2025stable} and DualCamCtrl~\cite{zhang2025dualcamctrl}. Both inject camera information using Plücker ray parameterization, where DualCamCtrl further incorporates monocular depth cues.
As shown in Table~\ref{tab:i2v_mix}, our method achieves the highest camera accuracy while maintaining comparable video fidelity. Beyond quantitative metrics, we provide qualitative comparisons on SDG-1.5M in \cref{fig:baseline_i2v}, as well as more results in \cref{fig:more_results_i2v} and \cref{fig:more_results_dv3dv}. While baseline methods are restricted to changing the viewpoint of a frozen scene, ReRoPE successfully generates dynamic subjects that adhere to the text prompt while following the specified camera trajectory.

\subsection{Ablation}
\label{sec:ablation}
We ablate key design choices of \method{} under the same I2V evaluation protocol as in \S\ref{sec:i2v}. 

\begin{figure}[t]
  \centering
  \includegraphics[width=\linewidth]{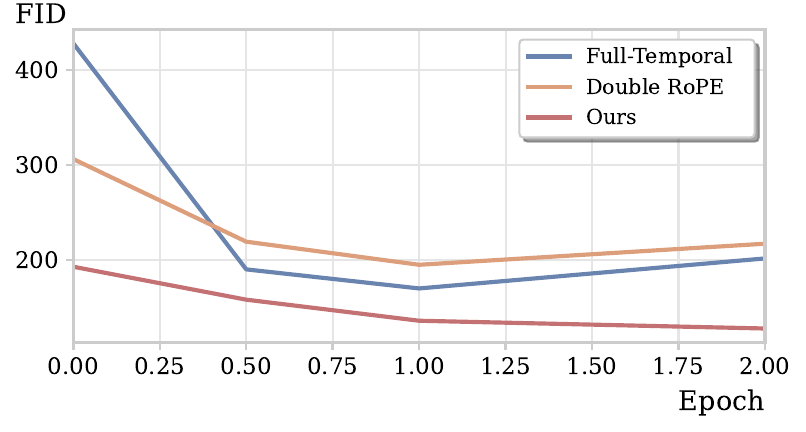}
  \vspace{-0.4cm}
  \caption{Training convergence comparison. FID scores over training epochs show that \method{} converges faster and reaches a significantly lower final FID than the Full-Temporal Replacement and Double RoPE baselines.    }
  \label{fig:ablation_curve}
  \vspace{-0.3cm}
\end{figure}

Beyond our proposed low-frequency injection, we consider two alternatives for integrating relative camera information into the pre-trained backbone: 1) Full-Temporal Replacement, which entirely replaces the temporal RoPE with the camera projection term, discarding original temporal frequencies; and 2) Double RoPE, which first applies the standard spatio-temporal RoPE from \cref{eq:rope-3d}, followed by an additional operation that replaces only the temporal component with the camera projection. This effectively rotates temporal features twice, first for positional encoding and then for camera transformation, while maintaining standard RoPE for spatial dimensions.
As shown in \cref{fig:ablation_curve}, our method achieves the best FID with the fastest convergence, demonstrating that it effectively preserves the pre-trained video generation prior. In contrast, Full-Temporal Replacement disrupts the backbone’s learned temporal prior, while Double RoPE forces the same channels to encode both temporal position and camera geometry. This creates signal interference that compromises both camera accuracy and perceptual quality. We provide a qualitative comparison in \cref{fig:ablation_rope_variations} and additional comparisons on more metrics in the supplementary material.

Second, we ablate translation normalization of the camera matrices. In this study, we compare \method{} with and without translation normalization, while keeping the same PRoPE injection strategy. Without normalization, large translation magnitudes in the projective block can inflate attention logits and destabilize conditioning, especially under larger baselines. Table~\ref{tab:ablate_tnorm} confirms this effect: enabling translation normalization consistently improves camera fidelity (lower RRE/RTE/ATE) and yields better FVD, indicating more stable and higher-quality generation under projective conditioning.

\begin{table}[t]
\centering
\small
\setlength{\tabcolsep}{4pt}
\renewcommand{\arraystretch}{1.15}
\begin{tabular}{lcccc}
\toprule
Setting & FID$\downarrow$ & FVD$\downarrow$ &
\makecell{Aesthetic\\Quality$\uparrow$} &
\makecell{Imaging\\Quality$\uparrow$} \\
\midrule 
w/o normalization & 
185.6123 & 110.8156 &
0.3495 & 0.4698 \\

\rowcolor{gray!12}
w/ normalization (Ours) & 
85.8568  & 59.4710 &
0.5671 & 0.6357 \\
\bottomrule
\end{tabular}
\caption{Ablation on translation normalization.}
\label{tab:ablate_tnorm}
\vspace{-0.5cm}
\end{table}

\section{Conclusion}
\label{sec:conclusion}
We recast camera control as a relative positional-encoding problem, leveraging the spectral redundancy inherent in pre-trained video transformers. By injecting camera geometry into the underutilized low-frequency temporal subspace of RoPE, \method{} provides a unified, plug-and-play framework applicable to both V2V and I2V tasks.
This approach preserves pre-trained generative priors, enabling precise control over complex trajectories without necessitating architectural changes. Our evaluations across diverse datasets demonstrate that \method{} achieves superior camera accuracy and 3D consistency while maintaining high visual fidelity.

\bibliographystyle{ACM-Reference-Format}
\bibliography{ref}
\clearpage

\section{Analysis of Low-Frequency Redundancy}
The choice of the base frequency $\theta=10^4$ in video diffusion models is a legacy from RoPE embeddings used in Large Language Models (LLMs)~\cite{su2021roformer, touvron2023llama}, which often model dependencies across tens of thousands of tokens and hence require a large base frequency. However, due to using factorized RoPE, video models typically operate on a much shorter temporal context during training ($T \approx 10$ to $128$ tokens).
In the temporal RoPE operator, the rotation angle increment per token for the $f$-th frequency band is given by $\Delta\phi_f = \omega_f = \theta^{-2f/d_\tau}$, where $\theta = 10^4$.
For a typical temporal head dimension $d_\tau = 32$, the lowest frequency band ($f=d_\tau/2 - 1 = 15$) yields a per-step rotation of $\Delta\phi_{15} \approx 10^{-4}$ rad. Even across a maximum temporal distance of $\Delta\tau = 100$ frames, the total phase accumulated is only $\mathcal{O}(10^{-2})$ rad, an amount so small that the rotation matrix is effectively the identity. This observation motivates \method~to repurpose this underutilized subspace as a carrier for camera signals.

\section{Data Annotation and Captioning Strategy}
\label{subsec:data_captioning}

To ensure high-fidelity alignment between textual descriptions and visual content, we utilized the \texttt{caption\_extend} script from the Wan2.2 pipeline for data preprocessing. Dense video captions were generated via the \texttt{Qwen3-VL Plus} API. 

To ensure that the camera control signal is driven exclusively by our ReRoPE embeddings rather than textual correlations, we implemented a prompt construction protocol designed to decouple visual content from camera motion:

    \textit{Decoupling Camera Motion:} We explicitly filtered out descriptors related to cinematic camera movements (e.g., ``pan left'', ``zoom in'') from the VLM's system prompt. This guarantees that the generated captions describe \textit{only} the scene content and object dynamics. Consequently, the diffusion model is forced to leverage the geometric signals provided by ReRoPE, preventing ``shortcut learning'' through textual cues.
    
    \textit{Multi-View Consistency:} For the MultiCam dataset, we enforced semantic consistency across viewpoints. Given that all videos within a single scene depict the same event, we generated a caption solely for the primary view and propagated this caption to all synchronized auxiliary views. This encourages the model to interpret varying visual inputs as geometric transformations of the same semantic content, rather than as content variations.
    
    \textit{Handling Static Scenes:} When processing static datasets, we modified the system prompt to exclude directives regarding dynamic motion analysis. This prevents the VLM from hallucinating motion in rigid scenes (e.g., describing a static building as ``moving''), thereby ensuring the text prompt accurately reflects the static nature of the environment.

\section{Pose Estimation and Scale Unification Protocol}
\label{subsec:pose_normalization}

In the Video-to-Video (V2V) inference pipeline, we employ VIPE~\cite{huang2025vipe} to extract the camera pose sequence from the source video. A primary challenge in this setting is the scale discrepancy between the estimated source trajectory (which possesses an arbitrary monocular scale) and the user-defined target trajectory.

To address this and ensure geometric consistency, we implement a robust two-stage normalization strategy:

    \textit{1) Individual Pre-normalization:} We first inspect the source and target trajectories independently to mitigate numerical instability caused by extreme scales. For a given trajectory $\mathcal{T} = \{t_i\}$, if the maximum translation magnitude exceeds a unit threshold ($\max_i \|t_i\|_2 > 1$), we normalize the entire sequence by its maximum norm. 
    
    \textit{2) Joint Normalization:} To unify the coordinate systems, we perform joint normalization on both trajectories. We define a global scaling factor $S$ based on the union of the pre-normalized source ($\mathcal{T}_{src}$) and target ($\mathcal{T}_{tgt}$) trajectories:
    \begin{equation}
        S = \max_{t \in \mathcal{T}_{src} \cup \mathcal{T}_{tgt}} \|t\|_2 + \epsilon
    \end{equation}
    Both trajectories are subsequently scaled by $S$. This ensures that the relative motion between the source and target views is preserved within a unified, model-friendly range ($[0, 1]$), effectively bridging the domain gap between distinct coordinate systems.

\section{Progressive Training Strategy}
\label{subsec:training_protocol}

To accelerate convergence, we employ a two-stage training strategy for the V2V model. We first train on short video clips ($T=10$) to efficiently learn the camera control injection. We then load this checkpoint and resume training on longer sequences ($T=42$). This progressive scaling significantly reduces the total training duration compared to training on long contexts from scratch.

\begin{table}[t]
\centering
\small
\setlength{\tabcolsep}{4pt}
\renewcommand{\arraystretch}{1.15}
\begin{tabular}{lcccc}
\toprule
Method & FID$\downarrow$ & FVD$\downarrow$ &
\makecell{Aesthetic\\Quality$\uparrow$} &
\makecell{Imaging\\Quality$\uparrow$} \\
\midrule
Full-Temporal Replacement & 
 135.4398 & 70.0219 &
0.5163 & 0.5757 \\

Double RoPE & 
118.7149 & 64.2370 &
0.5381 & 0.6070 \\

\rowcolor{gray!12}
ReRoPE (Ours) & 
85.8568 & 59.4710 &
0.5671 & 0.6357 \\
\bottomrule
\end{tabular}
\caption{Ablation on camera injection strategies.}
\label{tab:ablate_inject}
\end{table}

\section{Limitations}
\label{subsec:i2v_limitations}

While \method{} demonstrates robust 6-DoF control in static environments, its performance in Image-to-Video (I2V) settings is occasionally constrained by the inherent compositional bias of the web-scale datasets used for pre-training. Specifically, since most real-world videography tends to track salient subjects (e.g., humans) to maintain a central composition, the pre-trained backbone has acquired a strong inductive bias that couples camera movement with subject centering. This leads to an ``anchoring'' effect during inference: when a user-defined trajectory attempts to shift a subject away from the center or out of the frame entirely, the model's generative prior—which favors subject visibility—often counteracts the explicit camera control signal by synthesizing compensatory subject motion. We verify that this behavior stems from data-driven semantic priors rather than architectural flaws by evaluating \method{} on static scenes without clear subjects, such as the DL3DV dataset. In these scenarios, our model achieves precise responsiveness to all 6-DoF target poses, confirming that the control mechanism remains highly effective when it does not compete with strong subject-centering priors.

\begin{figure*}[t]
  \centering
  \includegraphics[width=\textwidth]{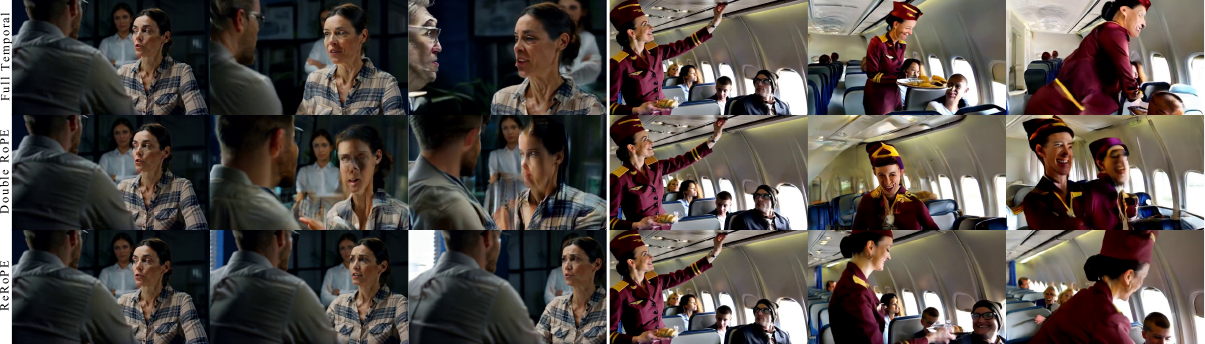}
  \caption{Comparison with ablation methods. }
  \Description{Qualitative comparison between ReRoPE and baseline methods showing video frames with different camera trajectories.}
  \label{fig:ablation_rope_variations}
\end{figure*}

\end{document}